\documentclass[applsci,article,accept,pdftex,moreauthors]{Definitions/mdpi} 
\usepackage[normalem]{ulem} % Enables strikethrough (\sout) and underlining

\firstpage{1} 
\pubvolume{1}
\issuenum{1}
\articlenumber{0}
\pubyear{2025}
\copyrightyear{2025}
\datereceived{ } 
\daterevised{ } % Comment out if no revised date
\dateaccepted{ } 
\datepublished{ } 
\hreflink{https://doi.org/} % If needed use \linebreak
\Title{Automatic Identification and Description of Jewelry through Computer Vision and Neural Networks for Translators and Interpreters}

\TitleCitation{Automatic Identification and Description of Jewelry through Computer Vision and Neural Networks for Translators and Interpreters}

\Author{José Manuel Alcalde-Llergo $^{1,2,}$*\orcidA{}, Aurora Ruiz-Mezcua $^{3}$\orcidC{}, Rocío Ávila-Ramírez $^{3}$\orcidD{}, Andrea Zingoni $^{2}$\orcidE{}, Juri Taborri $^{2}$\orcidF{} \mbox{and Enrique Yeguas-Bolívar} $^{1,}$*\orcidB{}}

\AuthorNames{José Manuel Alcalde-Llergo, Aurora Ruiz-Mezcua, Rocío Ávila-Ramírez, Andrea Zingoni and Enrique Yeguas-Bolívar}
\address{%
$^{1}$ \quad Department of Computer Science, University of Córdoba, 14071 {Córdoba}
, Spain\\
$^{2}$ \quad Economics, Engineering, Society and Business Organization, University of Tuscia, 01100 {Viterbo}, Italy 
\\
$^{3}$ \quad Department of Social Sciences, Translation and Interpreting, University of Córdoba, 14071 {Córdoba}, Spain\\}

\corres{Correspondence: jose.alcaldel@unitus.it (J.A.), eyeguas@uco.es (E.Y.)}

\abstract{Identifying jewelry pieces presents a significant challenge due to the wide range of styles and designs. Currently, precise descriptions are typically limited to industry experts. However, translators and interpreters often require a comprehensive understanding of these items. In this study, we introduce an innovative approach to automatically identify and describe jewelry using neural networks. This method enables translators and interpreters to quickly access accurate information, aiding in resolving queries and gaining essential knowledge about jewelry. Our model operates at three distinct levels of description, employing computer vision techniques and image captioning to emulate expert analysis of accessories. The key innovation involves generating natural language descriptions of jewelry across three hierarchical levels, capturing nuanced details of each piece. Different image captioning architectures are utilized to detect jewels in images and generate descriptions with varying levels of detail. To demonstrate the effectiveness of our approach in recognizing diverse types of jewelry, we assembled a comprehensive database of accessory images. The evaluation process involved comparing various image captioning architectures, focusing particularly on the encoder-decoder model, crucial for generating descriptive captions. After thorough evaluation, our final model achieved a captioning accuracy exceeding~90\%. }


\begin{document}

%%%%%%%%%%%%%%%%%%%%%%%%%%%%%%%%%%%%%%%%%%

\section{Introduction}

Image captioning presents a challenging task in the realms of computer vision and natural language processing, requiring the generation of textual descriptions for given input images~\cite{show_and_tell}. In the field of linguistics, particularly in translation and interpreting, this approach can provide accurate visual representations, precise technical terms, images, meanings, descriptions, and contexts. Translators and interpreters can rely on it to quickly access data about specific terms.

In recent years, deep learning-based approaches have emerged as a promising solution to the image captioning problem, harnessing the capabilities of deep neural networks to learn intricate mappings between images and text~\cite{survey}. Typically, an encoder-decoder structure is employed for this task, with neural networks serving as encoders to process the input image and extract high-level feature representations. Also, neural networks are commonly utilized as decoders, generating a sequence of words describing the image based on the features extracted by the encoder. The synergy between these two networks has significantly enhanced the performance of image captioning systems, approaching human-level performance on certain benchmark datasets~\cite{encoder-decoder}. 

An artificial intelligence (AI) tool capable of automatically identifying and describing jewelry can emulate human intuitive behavior, fostering increased connections between individuals from diverse cultures and customs. This includes overcoming language and cultural barriers by providing accurate and detailed descriptions of jewelry in various languages. Such an AI tool can encourage curiosity and cultural exchange by offering information about the history, symbolism, and craftsmanship of jewelry, sparking meaningful conversations and dialogues between individuals of different backgrounds. Additionally, it can promote appreciation and respect for diversity, fostering tolerance, inclusivity, and harmony among individuals and communities with different traditions and customs. Moreover, automatic jewelry recognition has practical applications in cultural commerce and tourism, assisting buyers and collectors in making informed decisions and engaging in fair trade of cultural jewelry, as well as attracting tourists interested in exploring and acquiring traditional jewelry during their travels.

In this study, various image captioning encoder-decoder structures will be assessed using a dataset comprising images of jewelry accessories. The dataset was created by collecting images of diverse accessories from two well-known online jewelry stores in the city of Córdoba, Spain. Córdoba is celebrated for its artisanal craftsmanship, particularly in the creation of high-quality silver jewelry. The city hosts notable jewelry fairs and events that exhibit distinctive designs, attracting both local and international buyers.

The utility of AI in linguistics becomes evident in unraveling the intricate linguistic landscape of jewelry terminology. AI-powered linguistic analysis can efficiently process vast datasets of historical documents, dowry letters, and cultural references related to jewelry. It helps identifying patterns, tracing linguistic evolution, and discerning the cultural contexts in which specific terms and expressions emerged.

Given the limited application of image captioning within this specific field of study, the task is intricate and notably captivating. The ultimate goal of the project is to generate detailed captions for accessories featured in the input images, highlighting characteristics such as their type (earrings, necklace, bracelet, etc.), material, color, and the specific jewelry they showcase. The images encompass individuals wearing the accessories, suggesting the potential application of the system to automatically generate descriptions for accessories worn by individuals in images sourced from fashion magazines, jewelry catalogs, and jewelry trade shows, among other sources. Additionally, the methodology enables the creation of simpler descriptions, indicating only the type of accessory, to facilitate a jewelry multiclass classification task. 

In the following sections, we navigate through key aspects of linguistic modeling, automatic jewelry recognition, and AI-based description generation. Section~\ref{sec:related-work} conducts a review of related work, highlighting prior research contributions. Section~\ref{sec:materials} delves into the linguistic levels and terminology specific to jewelry descriptions and explains the database and architectures supporting our approach. Section~\ref{sec:results} presents the experimental design and results, showcasing the effectiveness of our linguistic models, automatic recognition techniques, and AI-generated descriptions. Finally, in Section~\ref{sec:conclusions-future}, we draw conclusions and outline future research directions.

%%%%%%%%%%%%%%%%%%%%%%%%%%%%%%%%%%%%%%%%%%

\section{Related work}
\label{sec:related-work}
In this section, we conduct a thorough review of previous works that have significantly influenced and provided reference points for the current study. Our examination has revealed three main categories of research that are relevant to our investigation: the analysis of the impact of AI on linguistics, the application of artificial vision techniques for jewelry analysis, and neural networks for image captioning. Furthermore, we have undertaken a dedicated study specifically exploring the use of image captioning techniques in the context of jewelry; however, no prior research has been identified that specifically addresses this exact subject matter.

\subsection{AI and applied linguistics}

The confluence of AI and linguistics has cultivated a fertile ground for tackling specific challenges in the identification and description of jewelry in images. The capability of these integrated approaches to comprehend both visual content and natural language offers innovative solutions.

A pivotal theory influencing this domain is the hierarchy of grammatical complexity proposed in~\cite{Jackendoff2002,Jackendoff2016}. Their approach underscores that communication through natural language fundamentally involves mapping signals to meanings. The application of this theory extends beyond natural language and aligns with the goal of understanding and describing jewelry images.

Considering image processing, \citet{show_attend_and_tell} explore the connection between computer vision and natural language processing. The work emphasizes synergies between both fields for joint applications, such as generating descriptions of images. \citet{chen2016} take this approach a step further by introducing visual semantic role labeling, an advanced technique for image description generation, essential for detailed jewelry identification.

To undertake language modeling, a fundamental step for subsequent AI applications, studies such as those by~\cite{Manning2014} emphasize the need for robust language models.

Firstly, must be said that the distinction between translating from natural language into formal language and reasoning about that formal language was identified in~\cite{hillel:64,hillel:73}.

However, a long-standing goal of AI has been to build intelligent agents that can function with the linguistic dexterity of people, which involves such diverse capabilities as participating in a fluent conversation, using dialog to support task-oriented collaboration, and engaging in lifelong learning through processing speech and text~\cite{McShane2021}. 

These studies highlight the necessity of an interdisciplinary perspective, where linguistics and AI intertwine to advance jewelry image identification. They demonstrate how these converging disciplines can catalyze significant innovations in this domain.

\subsection{Computer vision for jewelry analysis}

Jewelry has played a pivotal role in human culture for millennia, and the capacity to distinguish various types of accessories holds significance across diverse fields, including security, e-commerce, and the analysis of social preferences. Recent years have witnessed substantial progress in computer vision models, particularly in the domains of object recognition and detection. Numerous models now exist that can proficiently discern a wide array of objects, ranging from animals and vehicles to accessories. An exemplar is DeepBE, introduced in~\cite{reco_jewels_face}, which adeptly categorizes accessories worn over the shoulders, such as necklaces, earrings, and glasses.

Despite these advancements, current methods for accessory recognition, like the aforementioned DeepBE, still exhibit certain limitations compared to human expertise. While these models effectively identify the presence of jewelry and other accessories in an image, they fall short in providing intricate details about the specific type of jewelry or its quality. Consequently, there arises a demand for more specialized models that can precisely and comprehensively recognize and classify jewelry and other accessories based on their unique characteristics.

Conversely, certain studies have been dedicated to crafting specialized models tailored to appraising the quality of specific types of jewelry. Notably, the research outlined in~\cite{pearls_quality} concentrates on the quality assessment of pearls based on images, employing computer vision techniques to identify and scrutinize the distinct visual characteristics of these items. The proposed methodology involves tracing light rays and scrutinizing highlight patterns to estimate the level of specularity within the input image of the pearl. This data is then employed to ascertain the equivalent index of appearance, serving as a metric for appraising the object's quality. Through the application of these techniques, the approach comprehensively evaluated the visual appearance of the object, offering valuable insights into its quality and potential industrial applications.

\subsection{Neural networks for image captioning}

Describing the content of an image was a task that, just a few years ago, could only be accomplished by humans. However, advancements in computer vision have led to the creation of new systems capable of replicating this human behavior with impressive performance.

The pioneering steps in this field were documented in works like~\cite{show_and_tell}, where the authors developed a probabilistic framework for generating descriptions from images. To achieve this goal, they devised an encoder-decoder structure employing neural networks. 

The primary function of the encoder is to process the input image, extracting meaningful features and encoding them into a format that can be easily understood by the neural network. In simpler terms, the encoder transforms the raw visual information of the image into a more abstract representation, highlighting important features. The decoder takes the encoded information from the encoder and generates a coherent and descriptive sequence. In the case of image captioning, this sequence comprises the natural language description of the image. The decoder essentially translates the encoded features back into human-readable language, allowing the system to generate meaningful and contextually relevant captions for the input images.

Neural networks come in various types, each designed for specific tasks. Two prominent types are Convolutional Neural Networks (CNNs) and Recurrent Neural Networks (RNNs).

CNNs are particularly well-suited for image-related tasks. They utilize layers of convolutional operations to automatically extract hierarchical features from input images. CNNs are effective in recognizing spatial patterns, making them commonly used in image classification, object detection, and image segmentation tasks.

On the other hand, RNNs are designed to handle sequential data and are well-suited for tasks involving a sequence of inputs, such as time-series data or natural language. RNNs have memory units that allow them to capture dependencies and relationships within sequential data. This makes them suitable for tasks like language modeling, speech recognition, and sequence generation.

Therefore, a good approach to addressing the problem of caption generation is the development of an encoder-decoder structure~\cite{Alcalde_Llergo_2023}. In this approach, a CNN is employed as the encoder, and a RNN serves as the decoder. This particular architecture has laid the groundwork for image captioning. In this process, the input image undergoes processing through the CNN to encode and extract features, which are then embedded into a fixed-length vector. This fixed-length vector, also known as an "embedding," succinctly represents the most important features of the image based on the encoding performed by the CNN. This vector representation is subsequently used in the process of generating descriptions (captions) for the image, where relevant information has been condensed into a structured and manageable format for the neural network. Choosing a fixed-length vector facilitates consistency and efficiency in processing information within the network. 

After the image has been processed through the CNN to encode features, the generation of final descriptions involves appending the RNN to the last hidden layer of the CNN. The term "last layer" in this context refers to the final or output layer of the CNN. In neural networks, layers are stacked, and the last layer typically produces the final representation or output. In image captioning, the information encoded by the CNN is transferred to the RNN at this stage to further refine and generate coherent textual descriptions.

The choice of utilizing a Long Short-Term Memory (LSTM) network as the RNN is significant. LSTM is a type of recurrent neural network designed to capture and retain long-term dependencies in sequential data. In the context of image captioning, LSTM plays a key role in understanding the sequential nature of language, allowing the model to generate descriptions that are not only contextually relevant but also exhibit a coherent flow. Its effectiveness lies in its ability to handle dependencies over longer sequences, making it well-suited for tasks involving the generation of sentences or captions.

Similarly, subsequently to the mentioned study, several works have emerged, incorporating innovative mechanisms to improve its outcomes. Particularly noteworthy is a proposal outlined in~\cite{show_attend_and_tell}, which introduced the concept of attention to the previously employed encoder-decoder architecture. In this scenario, diverse words generated by the LSTM were directed toward different sections of the input image. The outcomes demonstrated that their model produced highly accurate words that directly corresponded to the specific areas of the image that the LSTM focused on at each moment. As a result, a comprehensive and precise description of the image was achieved in the majority of cases.

Other advancements in image captioning have explored alternative approaches such as generative adversarial networks (GANs). A representative example is CgT-GAN, which combines adversarial training with CLIP-based semantic rewards to generate captions without relying on paired human annotations~\citep{Yu2023}. These methods can yield more diverse and creative outputs, especially in open-domain settings. However, GAN-based captioning models often suffer from training instability, require large computational resources, and may lack linguistic control, making them less suitable for tasks involving predefined syntactic structures like those used in our multi-level jewelry description framework~\citep{Shetty2017}.

Another line of research explores reinforcement learning (RL) techniques to enhance image captioning models. For instance, \citep{Chaffin2024} propose a strategy that integrates ground truth captions into a CLIP-guided RL framework. This approach aims to generate more distinctive and fluent captions by leveraging the similarity between generated captions and corresponding images. Similarly, in~\citep{Shi2018} authors introduce a RL approach that combines a policy network, responsible for generating captions, with a value network that estimates the expected reward of a generated sequence. This architecture is optimized using sequence-level feedback, allowing the model to refine caption quality based on global evaluation metrics rather than word-by-word accuracy. While RL-based models have demonstrated improvements in generating diverse and contextually relevant captions, they often require extensive training data, careful reward design, and complex tuning processes~\citep{Ghandi2023}.

%%%%%%%%%%%%%%%%%%%%%%%%%%%%%%%%%%%%%%%%%%
\section{Materials and Methods}
\label{sec:materials}

This section outlines the methodological framework adopted to develop a system capable of generating descriptive captions for jewelry images. The proposed approach integrates insights from linguistics and computer vision to address the complexity of jewelry representation. Throughout the following subsections, we describe the linguistic foundations that support the structuring of jewelry descriptions, the creation and preprocessing of the image dataset, and the experimental setup used to train and evaluate the captioning models.

\subsection{Levels and terminology for the description of jewelry}

Our primary objective is to create a concise and cohesive summary of jewelry images. However, grasping the high-level informational content of these images poses challenges. Initially, a jewelry image conveys a wealth of information, making it impractical to provide all details textually. Secondly, the choice of a jewelry image as a communication medium allows viewers to extract information at various levels. A cursory glance may convey the intended message and salient features, while a more in-depth examination offers further insights.

% \subsection{Levels of description}
% \label{sec:levels}

To address the task of identifying content for jewelry image summaries, we leverage insights gained from an experiment where online jewelry catalogs were used to generate summaries for various images with a shared high-level intention. We follow a similar approach to that employed for graphic descriptions in~\cite{McCoy2001}. The key finding was that the intended message of a jewelry image was consistently conveyed in all summaries, regardless of visual features. Online catalogs, emulating human behavior, enhanced the intended message with salient features, depending on the image's purpose. Similar summaries for a specific jewelry image led us to hypothesize shared perceptions of salient features. Although the set of salient features may be consistent for images with the same underlying intention, differences in summaries arise from the saliency of these features. The exclusion of exhaustive details in summaries aligns with Grice's Maxim of Quantity~\cite{grice1975}, emphasizing informative contributions relevant to the current exchange.

To identify specific content in jewelry images, we will focus on a set of propositions capturing meaningful information. This set encompasses a variety of details present in the images, including propositions common to all images and propositions specific to certain message categories. Some of these propositions include: labels for all jewelry items, type of jewelry, materials used in the jewelry, relationship between the materials, colors present in the jewelry, style or design of the jewelry, specific details like engraved patterns, size or relative dimensions of the jewelry and overall impression or aesthetic, including the addition of descriptive adjectives. These propositions cover a spectrum of information, allowing for a comprehensive and nuanced summary of jewelry images.

From a linguistic perspective, the justification for developing linguistic models based on different theories that propose levels of descriptions~\cite{dijk77, leech92, mairal09, abdel21} for jewelry is rooted in the need to capture and express the semantic diversity and inherent complexity of jewelry pieces~\cite{vaswani2017,Bender2021}. Each level of description has its own linguistic characteristics that enable effective communication of specific information about  jewelry. Here we provide the descriptions considered in our approach and specific justifications for each level of description:

\begin{itemize}
    \item \textbf{Basic Description: Noun + Noun.}
    \begin{itemize}
        \item \textit{Communicative efficiency:} This simple format focuses on the essence of the jewelry by combining two key nouns. It provides concise and direct information about the type of jewelry present. In contexts where brevity is essential, this format allows for quick identification of the jewelry type.
        \item \textit{Clarity and simplicity:} Ideal for quickly identifying the type of jewelry without adding unnecessary details. The straightforward structure enhances clarity, making it suitable for scenarios where a quick overview is required.
    \end{itemize}
    \item \textbf{Normal Description: Adjective + Noun + Adjective + Noun.}
    \begin{itemize}
        \item \textit{Semantic expansion:} The inclusion of adjectives allows for a more detailed and nuanced description. It enables conveying specific features such as color, material, or design. This format provides a broader semantic scope, accommodating a richer set of details.
        \item \textit{Aesthetic appeal:} By adding adjectives, the beauty and visual attributes of the jewelry are highlighted. This can be relevant in contexts where aesthetics plays a significant role, such as art exhibitions or design showcases.
    \end{itemize}
    \item \textbf{Complete Description: Superlative Adjectives + Noun + Complement.}
    \begin{itemize}
        \item \textit{Comprehensiveness and specificity:} The inclusion of superlative adjectives indicates the exceptional quality of the jewelry, offering a complete and detailed description. This format is essential for describing unique or high-value pieces, providing comprehensive insights.
        \item \textit{Differentiation and valuation:} Allows highlighting the unique features that make the jewelry stand out among others. Especially valuable in contexts where exclusivity is sought, such as luxury markets or bespoke jewelry presentations.
    \end{itemize}
\end{itemize}

The development of linguistic models that encompass these types of descriptions allows for a more comprehensive and accurate representation of jewelry. Such models cater to the variety of information that may be relevant in different situations and contexts, ensuring that linguistic descriptions effectively convey the richness and uniqueness of each jewelry piece.

Using this approach, we propose some examples of descriptions at different complexity levels after recognizing the jewelry pieces (see Table~\ref{tab:jewellery_descriptions}). The presence of superlative adjectives is always optional, and a complete description can exist without them. For example, "Earrings in sustainable yellow gold adorned with exquisite, brilliant-cut diamonds and featuring a secure push-back clasp" is equivalent to "Earrings in yellow gold with brilliant-cut diamonds and a push-back clasp."

\begin{table}[ht]
\centering
\caption{Description levels of jewelry pieces}
\label{tab:jewellery_descriptions}
{\footnotesize
\begin{tabularx}{\textwidth}{@{}lXXX@{}}
\toprule
\textbf{Level} & \includegraphics[height=2cm]{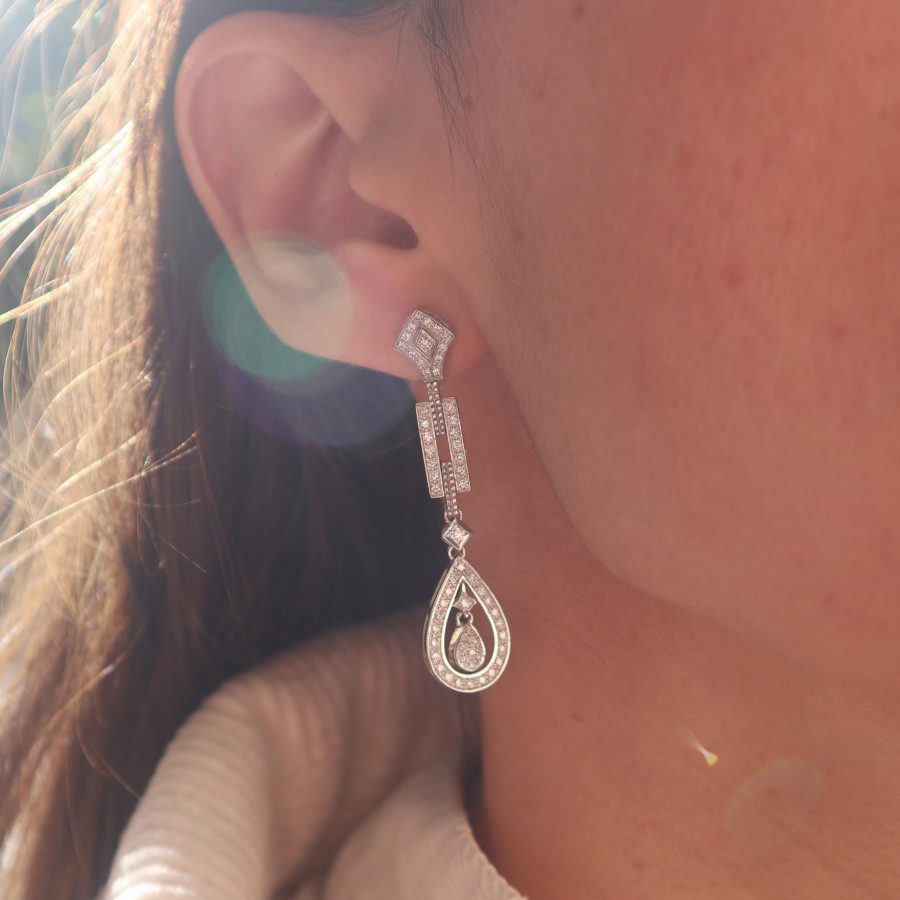} & \includegraphics[height=2cm]{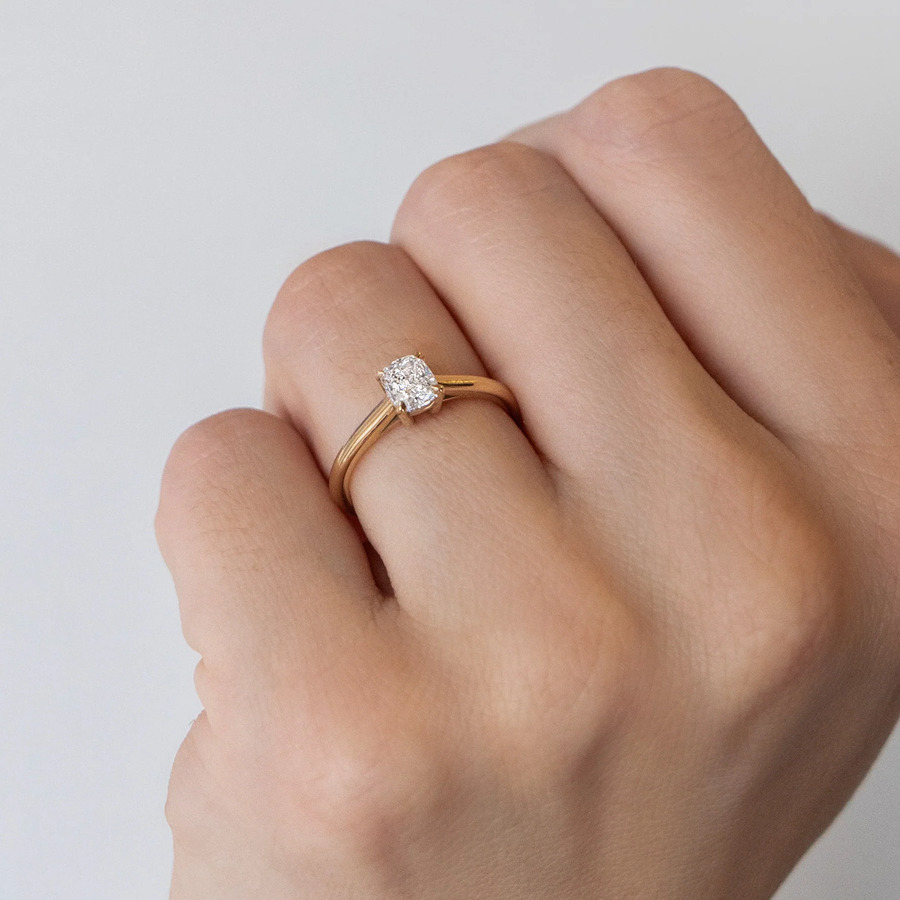} & \includegraphics[height=2cm]{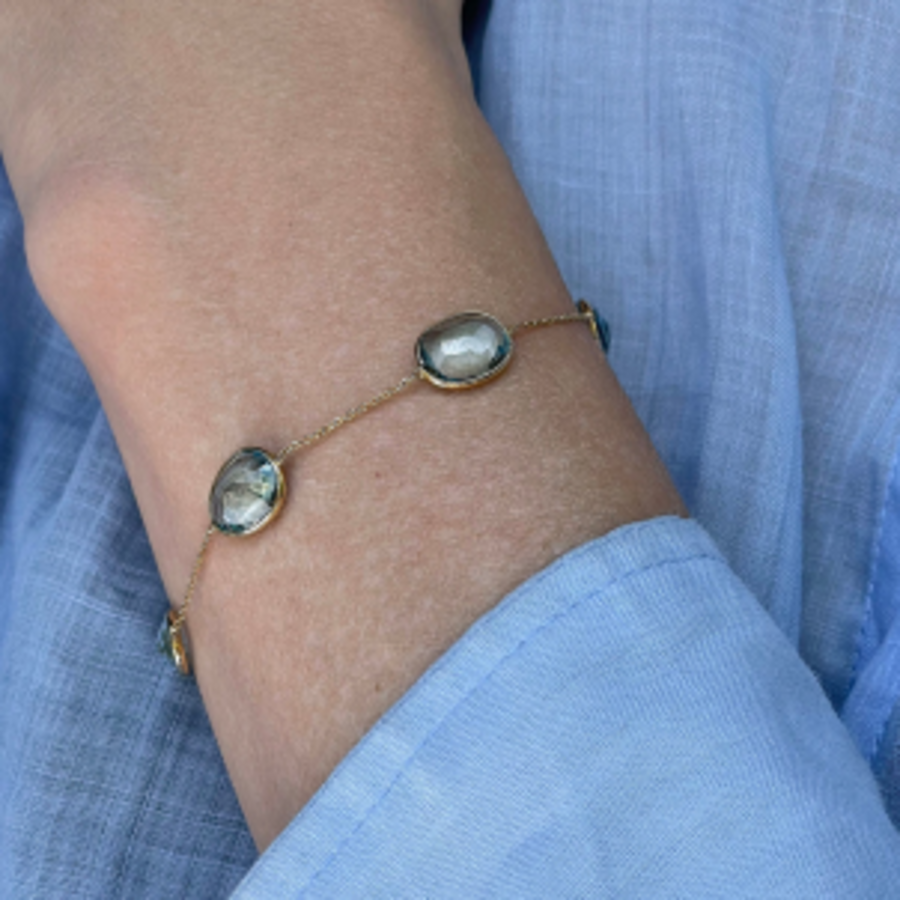} \\
\midrule
\textbf{Basic}    & Earrings in yellow gold. & Solitaire in rose gold. & Yellow gold bracelet. \\
\textbf{Normal}   & Yellow gold and diamond earrings. & Ring in rose gold with central diamond. & Yellow gold bracelet with topazes. \\
\textbf{Complete} & Earrings in sustainable yellow gold adorned with exquisite, brilliant-cut diamonds and featuring a secure push-back clasp. & Iris solitaire in rose gold with central diamond. & Sustainable yellow gold bracelet adorned with dazzling sky topazes. \\
\bottomrule
\end{tabularx}
}
\end{table}

At the end, systematizing jewelry terms, determining their role, functional, and semantic meaning would help to offer high-quality translations by AI. In the transition to a modern digital system, it is relevant to provide background knowledge that contributes to understanding the meaning of concepts and to shift towards new electronic forms of creating dictionaries, glossaries, and thesauri for high-quality translation~\cite{Kozhakhmetova}.

\subsection{Fundamentals of jewelry terminology}

In all languages a lot of idiosyncrasy and cultural referents can be found, especially when we work with specialized texts and speeches. Each jargon is based on the culture where it was originally used or created. This especially happens when a sector is prominent in a society. For instance, wine vocabulary is extensive and rich in most Mediterranean countries, but not that much in regions like Alaska, where the weather and based-habits are completely different from wine areas. In the case of jewelry, we can find a wide range of specific terminology that, at times, is particularly difficult to translate, for example, from Spanish (being Spain a traditional power in the field) into other languages.

Terminology related to jewelry is very wide and varied. It ranges from numerous nouns to name each piece of jewelry to an infinite number of adjectives describing its shape, cut, weight (carat), color or materials it is made of. It can therefore be said that the descriptions of jewelry items are very specific, exhaustive, and dominated by the use of adjectives. There are also different varieties and terms depending on geographical areas and idiolects, as well as cultural implications that underlie jewelry terminology. Nevertheless, there are not many reliable resources from a linguistic, lexicographical and a translational point of view. Most works focus on one hand, on Chemistry, especially on composition, typology and materials's alloy \cite{demortier09,yu18} and, on the other hand,  on Commerce and Marketing \cite{ariza06, arellano09, diaz14, hani18, hernandez18}. Considering the growing demand for internationalization in this sector, there is a strong need to create resources which would help professionals to standardize and use the accurate terminology jewelry.

In the context of jewelry, the terminology associated with descriptive words is rich and diverse. Similar to languages that lack a distinct class of adjectives, the descriptions of jewelry items often involve the use of nouns and verbs to convey specific characteristics. Additionally, the concept of gender or noun class, which is common in language, is mirrored in jewelry terminology. For instance, certain jewelry pieces may be classified into categories based on gender, such as masculine or feminine designs, influencing how they are described and categorized. This linguistic parallel illustrates the intricate and nuanced nature of expressing the features of jewelry items.

In the scope of linguistic descriptions, the clarity of explanations often follows the principle that the better the descriptions, the more evident they become, particularly given that those reading them possess a functional knowledge of language. This characteristic holds true for various scientific fields, where intricate issues are articulated in detailed descriptions.

In the context of this research, AI plays a key role in object recognition. The computer vision system identifies objects, and linguistically, we assign a noun to the object along with a brief description. Essentially, jewelry-specific terminology and adjectives are utilized in this linguistic process \cite{Liddicoat2004}.

The catalog or vocabulary list employed in our approach could provide a broad array of possibilities, but they generally fall into larger categories: fobs, lockets, necklaces, bracelets, armlets, finger rings, watches, buttons, clips, cufflinks, pendants, pins, tie clasps, anklets, toe rings, ear ornaments, hair ornaments, and nose ornaments. Furthermore, each group encompasses various items.

The size and shape should be supplemented with details about the material it is made of, the color, and other specifications such as weight and care. Strictly speaking, there are only seven precious stones: pearl, diamond, ruby, emerald, alexandrite, sapphire, and oriental catseye. In contrast, the so-called semi-precious stones include amethyst, topaz, tourmaline, aquamarine, chrysoprase, peridot, opal, zircon, and jade. %Information about their hardness, color, and sources of supply is provided.

Information about the hardness, color, and sources of supply of a specific gemstone can be found in various places, and the availability of information may depend on the type of gem and its supplier. Some common places where you might find this information are:
gem and diamond certificates, product descriptions in jewelry stores, websites of jewelry suppliers and designers, and books and online resources on gemmology.

In our case, we have relied on a set of catalogs provided by the \href{https://parquejoyero.es/en/home}{Parque Joyero de Córdoba}. Using these catalogs, we have created a relational database encompassing materials, precious and semi-precious stones, commonly used terms and tokens, most common meanings, colors, shapes, and common interrelationships in the creation and presentation of each piece of jewelry.

The database serves as a comprehensive resource, capturing a wide array of information related to the field of jewelry. It includes details about the materials used, the types of precious and semi-precious stones featured, prevalent terminology and expressions, as well as the common meanings associated with various elements. Additionally, the database covers information about colors, shapes, and the typical interconnections found in the crafting and display of each jewelry item. This extensive compilation of data provides a foundation for our linguistic models and aids in the development of a nuanced and contextually rich understanding of jewelry descriptions.

\subsection{Creation of the dataset}
The jewelry dataset used in the experiments was specifically crafted for this study by extracting, preparing, and merging images from two online jewelry stores\footnote{Baquerizo Joyeros: \href{https://baquerizojoyeros.com}{https://baquerizojoyeros.com}}\footnote{Doñasol: \href{https://doñasol.com}{https://doñasol.com}} in Córdoba (Spain). These catalogs included multiple images of the same jewelry item, captured from different perspectives or highlighting particular details (e.g., side views, clasps, gemstone close-ups). These naturally diverse viewpoints were preserved and included in the dataset, enriching the training data beyond standard augmentation techniques.

Nevertheless, as is common in computer vision challenges, the available quantity of images was not sufficient to train robust models adequately for optimal results. To address this limitation, various data augmentation techniques were implemented to diversify the dataset. The augmentation techniques applied to enhance the jewelry database include: 90º rotations, width shift by 30\%, length shift by 30\%, cuts by 15\%, enlargement or reduction by 5\%, slight color changes, horizontal and vertical flips and brightness range by 80\%.

\begin{figure}
        \centering
        \includegraphics[width=0.95\textwidth]{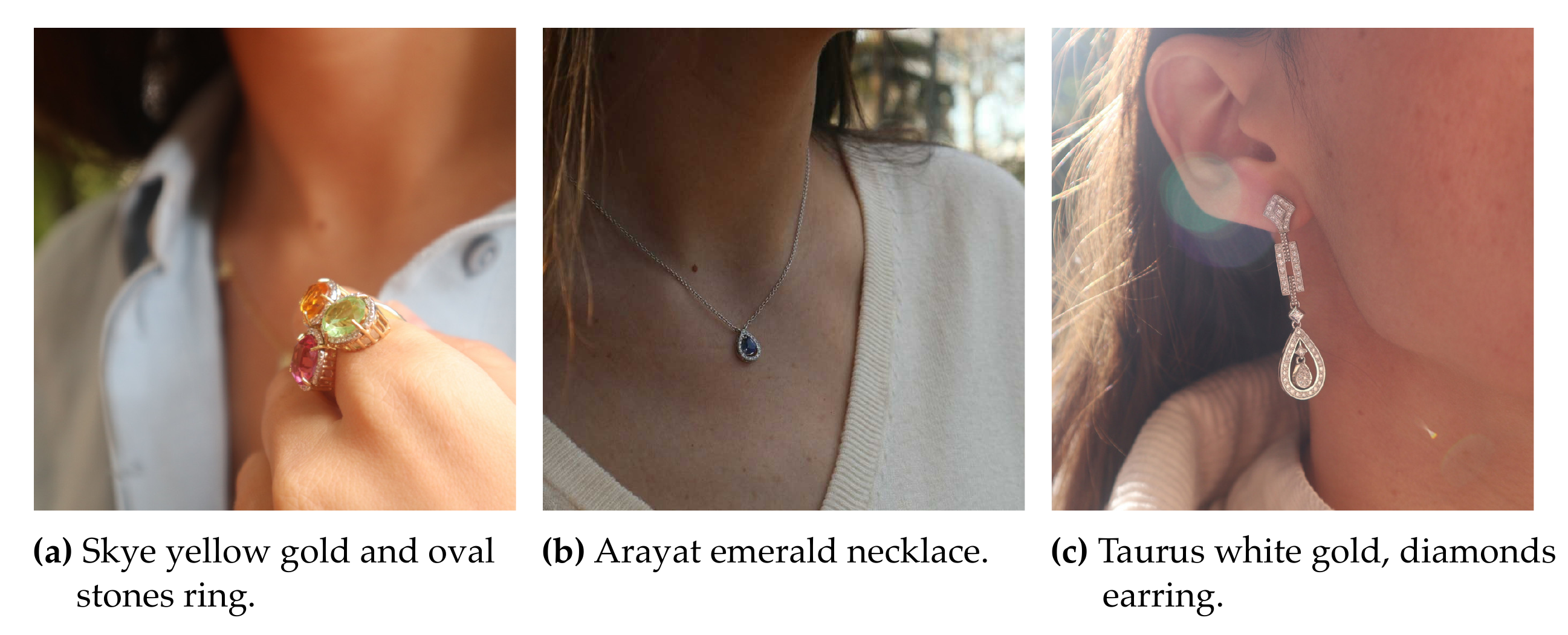}
        \caption{Training images and captions from the final dataset.}
        \label{fig:joyas}
    \end{figure}

Following the data augmentation procedures, the merging process resulted in a comprehensive and unified jewelry database comprising a total of 5374 accessory images. For experimentation, this dataset was partitioned into training (75\%), validation (15\%), and testing (10\%) sets. Figure \ref{fig:joyas} displays some samples from the dataset along with their corresponding captions.

\subsection{Experimental design}

\begin{figure}[h]
    \centering
    \includegraphics[width=0.99\linewidth]{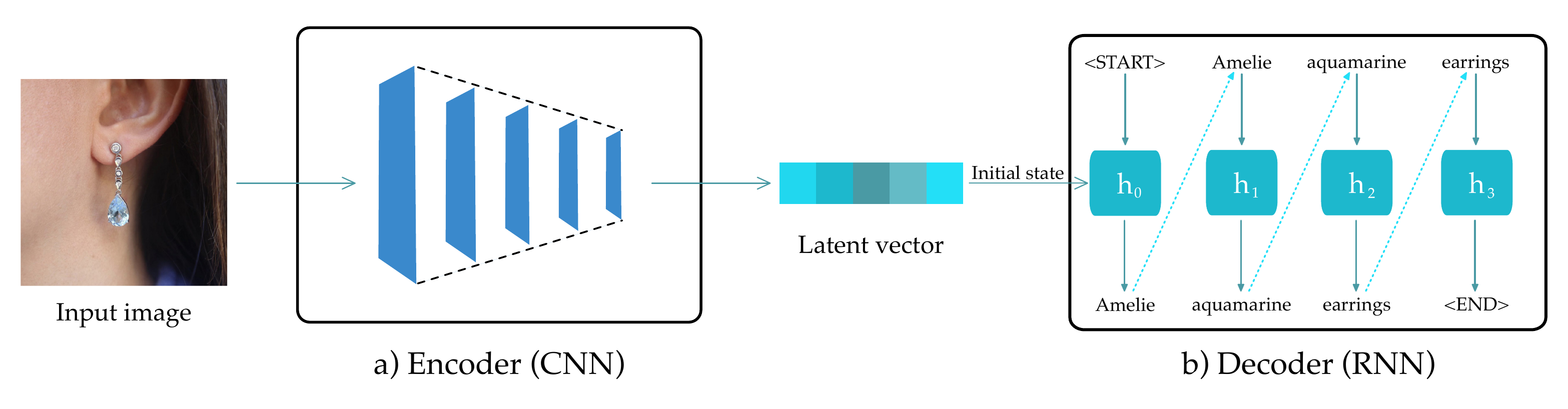}
    \caption{Overview of the encoder–decoder model used for jewelry description. The CNN encoder~(a) processes the input image of a jewelry item into a latent feature vector, 
    which is then used to initialize the initial state of the RNN decoder (b) that generates the caption.}
    \label{fig:overview}
\end{figure}

The first step in designing the experimentation was to select models that could be relevant for addressing the task of jewelry description. The presence of pre-trained neural networks, enriched with extensive datasets, allows us to leverage their capabilities for enhanced results. Consequently, there is no need to build a neural network from the ground up. This approach of repurposing networks is known as Transfer Learning \cite{transferLearning}. In the experimentation phase, various combinations of CNNs and RNNs architectures were explored to identify the optimal encoding and decoding structure for describing the accessory images within the database. Figure~\ref{fig:overview} provides an overview of the proposed architecture for generating a Basic-level jewelry description.

The encoder-decoder structure is selected due to its proven effectiveness in tasks involving short, structured textual outputs, such as image captioning~\citep{show_and_tell}, and considering the nature of the available dataset. In our case, the generated textual outputs follow predefined linguistic templates (basic, normal, and complete levels), which significantly reduces the complexity of the sequence modeling task. Furthermore, this architecture offers a favorable trade-off between interpretability, computational efficiency, and performance, especially in low resource settings where training data and processing capabilities are limited. Although recent Transformer-based architectures~\citep{captioningTransformers} have achieved state-of-the-art results in large-scale image captioning tasks, they typically require significantly higher computational resources and larger datasets. For these reasons, the CNN–RNN model remains a solid and accessible solution for domain-specific applications that prioritize structured linguistic output and lightweight deployment.

Three prominent architectures have been explored for CNNs, each with its unique characteristics: VGG-16 \cite{vgg16}, InceptionV3 \cite{inceptionv3} and MobileNet \cite{mobilenet}.

On the other hand, in the field of sequence modeling and predictions, RNNs have been instrumental. However, a common challenge faced by traditional RNNs is short-term memory loss, which hampers their ability to capture long-term dependencies within the data. To overcome this limitation, advanced models like Long Short-Term Memory (LSTM), see \cite{lstm}, and Gated Recurrent Unit (GRU), see \cite{gru}, have been introduced.
Both LSTM and GRU play pivotal roles in enhancing the linguistic finesse of the proposed architectures, making them proficient in describing various jewelry types in images with linguistic sensitivity.

In the context of experimental design, considerations extended beyond the choice of CNN architecture and RNN type. Several critical parameters were systematically explored:

\begin{itemize}
    \item \textbf{Number of neurons in the RNN:} This parameter affects the model's capacity to capture complex linguistic patterns. The values considered were 64, 128, 256, 512, and 1024.

    \item \textbf{Batch size:} This determines the number of samples processed per training iteration. Tested batch sizes included 4, 8, 32, 128, and 512.

    \item \textbf{Learning rate:} This regulates the step size during the optimization process and is essential for effective model convergence. The explored values were 0.0001, 0.001, 0.01, and 0.1.

    \item \textbf{Optimizer:} Optimizers influence parameter updates during training, affecting convergence speed and stability. The following optimizers were evaluated: Adam, Adagrad, Adadelta, and RMSProp.
\end{itemize}

To reduce the risk of overfitting, an early stopping mechanism that monitors the validation loss is implemented. When no improvement is observed over a certain number of epochs, training is halted to prevent the model from fitting non-generalizable patterns in the training data.

Regarding all these considerations, carefully designed experiments were conducted to enable a robust comparison of different parameter configurations for model generation. Following these experiments, the optimal model configuration for generating captions from the jewelry database is determined.

The evaluation of caption quality poses challenges due to the intricate nature of image captioning. While established metrics like METEOR, BLEU, or ROUGE \cite{metrics} exist for measuring language and semantic accuracy, our dataset's unique constraints led us to adopt a straightforward approach. We deemed generated captions correct if they matched the original ones provided in the online jewelry stores. For future implementations, there is an intention to enhance the dataset and incorporate these established metrics for a more comprehensive assessment of our model's performance.

%%%%%%%%%%%%%%%%%%%%%%%%%%%%%%%%%%%%%%%%%%
\section{Results and discussion}\label{sec:results}
In this section, we present the results corresponding to the best-performing configurations obtained after a comprehensive evaluation of all tested model architectures and hyperparameter settings. Initially, a classification task was undertaken, categorizing accessories into four distinct classes: necklaces, rings, earrings, and bracelets. After thorough analysis to determine the optimal configuration for each considered encoder and decoder combination, the most favorable outcomes in terms of test Correct Classification Rate (CCR) were attained using VGG-16 as the CNN and GRU as the RNN, as detailed in Table~\ref{tab:model_comparison}. This table also includes additional metrics, such as validation CCR and Loss.

\begin{table}[h]
\centering
\caption{Best model for identification of the class: necklace, ring, earring, and bracelet.}
\label{tab:model_comparison}
\resizebox{0.8\textwidth}{!}{% Adjusts the table to text width
\begin{tabular}{cccccc}
\toprule
CNN & RNN & Neurons & Val. CCR & Val. Loss & Test CCR \\
\midrule
InceptionV3 & GRU & 1024 & 0.9187 & 0.1026 & 0.9161 \\
\textbf{VGG-16} & \textbf{GRU} & \textbf{512} & \textbf{0.9314} & \textbf{0.0390} & \textbf{0.9464} \\
MobileNet & GRU & 64 & 0.8851 & 0.2696 & 0.8802 \\
InceptionV3 & LSTM & 1024 & 0.9487 & 0.0301 & 0.9193 \\
VGG-16 & LSTM & 512 & 0.9597 & 0.0542 & 0.9227 \\
MobileNet & LSTM & 64 & 0.8811 & 0.2208 & 0.8775 \\
\bottomrule
\end{tabular}
}
\end{table}

Following the selection of the best configuration, a more comprehensive analysis was conducted, incorporating additional metrics like precision, recall, and F1-Score to assess the model's performance in classifying each jewelry class independently. Precision measures the accuracy of positive predictions, recall assesses the model's ability to capture all positive instances, and F1-Score provides a balance between precision and recall, offering a comprehensive evaluation of the model's performance. Table \ref{tab:ind_classification} illustrates the model's effectiveness in the classification task, achieving scores exceeding 91\% for all metrics and accessory types, with the exception of bracelets, which proved more challenging to detect.

\begin{table}[h]
\centering
\caption{Individual class classification quality.}
\label{tab:ind_classification}
\resizebox{0.5\textwidth}{!}{% Adjusts the table to text width
\begin{tabular}{cccccc}
\toprule
Accessory & Precision & Recall & F1-Score \\
\midrule
Necklaces & 0,9452 & 0,9087 & 0,9131 \\
Rings & 0,9276 & 0,9173 & 0,9343 \\
Earrings & 0,9452 & 0,9675 & 0,9674 \\
Bracelets & 0,9420 & 0,8298 & 0,8807 \\
\bottomrule
\end{tabular}
}
\end{table}

It's worth noting that an optimization of the various parameters under consideration has been conducted, yielding the best results with the following configuration: number of neurons = 512; batch size = 8; optimizer = Adam and learning rate = 0.001.

In the context of the primary task of generating comprehensive captions from images of accessories, optimal results were achieved with the VGG-16 and GRU combination. MobileNet was excluded from this investigation due to its tendency to frequently get stuck in local optima, leading to suboptimal performance in handling the specific challenges posed by image captioning in the context of jewelry. Table \ref{tab:best_model} presents a comparison of all considered encoder-decoder combinations for this task.

\begin{table}[h]
\centering
\caption{Best model for image captioning.}
\label{tab:best_model}
\resizebox{0.8\textwidth}{!}{% Adjusts the table to text width
\begin{tabular}{cccccc}
\toprule
CNN & RNN & Neurons & Val. CCR & Val. Loss & Test CCR \\
\midrule
InceptionV3 & LSTM & 512 & 0.9307 & 0.0615 & 0.8439 \\
VGG-16 & LSTM & 512 & 0.9270 & 0.0881 & 0.9036 \\
InceptionV3 & GRU & 256 & 0.9483 & 0.0921 & 0.8354 \\
\textbf{VGG-16} & \textbf{GRU} & \textbf{256} & \textbf{0.9633} & \textbf{0.0706} & \textbf{0.9345} \\
\bottomrule
\end{tabular}
}
\end{table}

In this experimentation, an optimization of the different parameters considered has been carried out, resulting in the best outcomes with the following configuration: number of neurons = 256; batch size = 16; optimizer = Adam and learning rate = 0.001.

Furthermore, the assessment of our proposed approach revealed that the top-performing model exhibited excellent captioning accuracy, closely aligning with the original labels from online jewelry stores. Notably, the captioning accuracy for rings, necklaces, and earrings are very promising. However, the classification accuracy for bracelets, while still commendable, was slightly lower in the test samples. The evaluation criteria involved a meticulous comparison of whether the generated captions precisely matched the authentic descriptions of the accessories captured in the images.

It's important to acknowledge certain limitations in our model. The observed misclassifications, particularly in the case of bracelets, can be attributed to the inherent challenges posed by accessories with similar shapes and materials. Additionally, instances arose where the same type of accessory could exhibit variations in materials or jewels, further contributing to the nuanced nature of accurate classification. These challenges highlight areas for potential refinement and improvement in future iterations of the model.

\begin{figure}[h]
    \centering
    \includegraphics[width=0.6\textwidth]{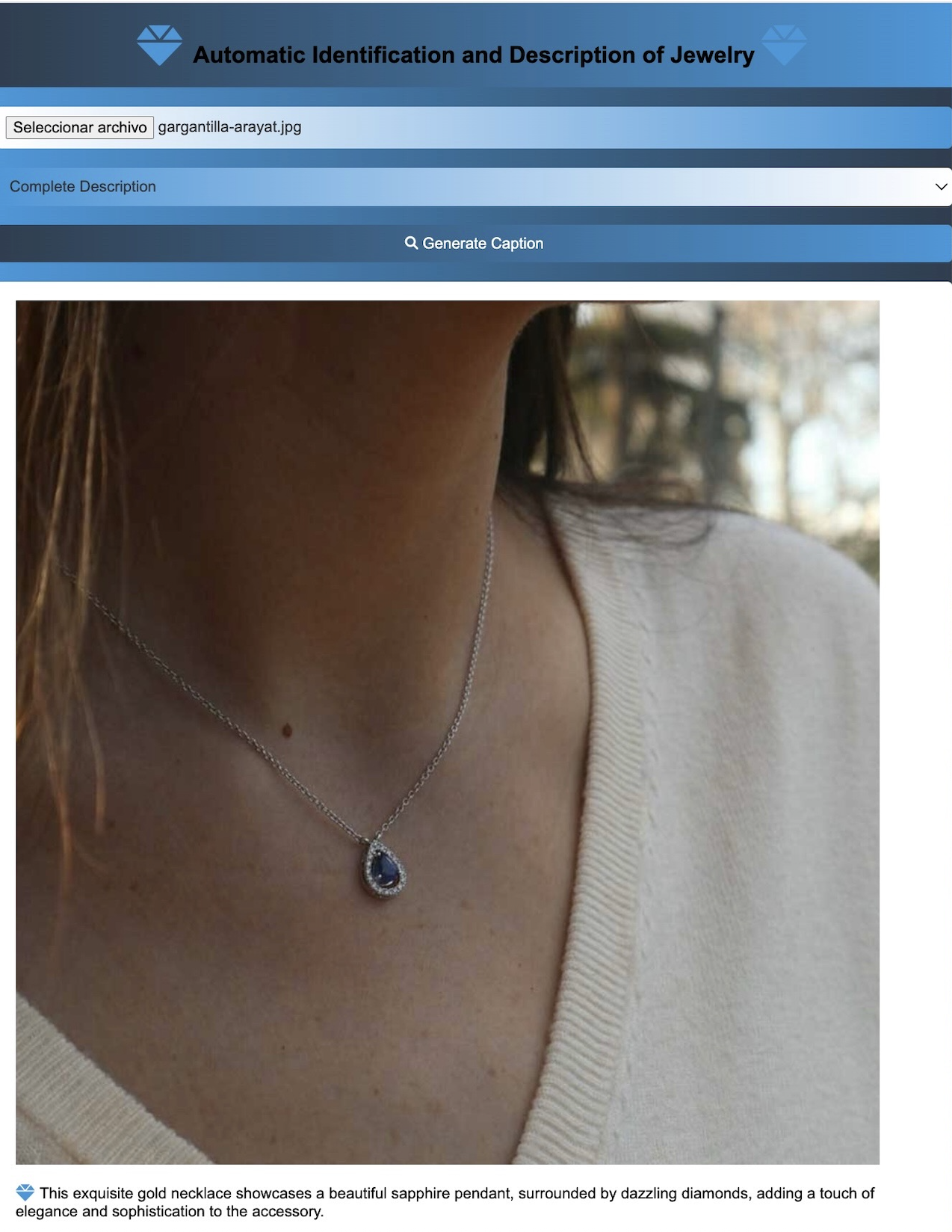}
    \caption{Jewelry image captioning interface (complete description after several iterations for generating the caption).}
    \label{fig:web}
\end{figure}

Ultimately, a three-tiered interface has been developed, leveraging the best-performing image captioning models created. This interface allows users to upload an image of an accessory and generate a corresponding description. Illustrated in Figure \ref{fig:web}, the interface offers descriptions of varying complexity (see Section~\ref{sec:materials}).

Let's consider an example image of a gold necklace with a pendant featuring a sapphire gemstone. The image is uploaded to the system, and the generated descriptions at different complexity levels would be as follows:

\begin{itemize}
    \item \textbf{Basic Description:} ‘A necklace with a pendant.’
    \item \textbf{Normal Description:} ‘A gold necklace with a sapphire pendant.’
    \item \textbf{Complete Description (after several iterations):} ‘This exquisite gold necklace showcases a beautiful sapphire pendant, adding a touch of elegance and sophistication to the accessory.’
\end{itemize}

From a linguistic perspective, the descriptions exhibit a progressive level of detail, ranging from a simple identification of the accessory to a more elaborate narrative that captures specific attributes such as material, gemstone, and an emotional tone. This iterative process in the complete description phase involves multiple iterations to incorporate the embellishment of the jewelry with superlative adjectives, further enhancing the richness of the generated descriptions. This meticulous approach aims to provide potential customers with a more nuanced and compelling understanding of the featured jewelry.

For the sake of reproducibility, both the jewelry identification models, multi-level description generation models, and the database of jewelry images, along with the necessary code, are available at 
\href{https://github.com/jewelryling/jewelry\_linguistics/}{Jewelry Linguistics Github}.

%%%%%%%%%%%%%%%%%%%%%%%%%%%%%%%%%%%%%%%%%%
\section{Conclusions and Future work}
\label{sec:conclusions-future}

The encoder-decoder architecture, coupled with linguistic models, has proven to be instrumental in generating coherent and contextually relevant descriptions across different types of jewelry. This multilevel linguistic approach contributes to a more versatile and accurate representation of jewelry items. It allows for a tailored and informative description that aligns with diverse user preferences and contextual requirements. The adaptability and scalability of our model underscore its potential in the future for broader applications in the domain of automated image captioning, with implications for diverse industries beyond jewelry analysis.

In considering future avenues for research, the refinement of our linguistic models stands out as a priority to enhance accuracy and accommodate an even broader spectrum of jewelry variations. Exploring advanced network architectures and expanding the dataset to encompass diverse cultural and stylistic influences can contribute significantly to the development of a more robust and culturally sensitive linguistic model. In particular, incorporating attention mechanisms into the encoder-decoder architecture could improve the model’s ability to dynamically focus on relevant visual features, helping to mitigate repetition and enrich caption diversity. Building upon this, we also plan to explore Transformer-based architectures, which leverage self-attention to model long-range dependencies and enable parallel generation, making them especially suitable for producing the complex and expressive descriptions targeted in this work.

Another important direction for future work is to evaluate the robustness of the system under real-world image acquisition conditions, including varying lighting, reflections, and occlusions, which are common in user-generated content and may affect the accuracy of jewelry recognition and description.

Furthermore, incorporating user feedback and preferences into the training process will be pivotal in refining the models for personalized and context-aware jewelry descriptions. This user-centric approach aims to ensure that the generated descriptions align closely with individual preferences and contribute to a more user-friendly and adaptable system.

%%%%%%%%%%%%%%%%%%%%%%%%%%%%%%%%%%%%%%%%%%
\vspace{6pt} 

%%%%%%%%%%%%%%%%%%%%%%%%%%%%%%%%%%%%%%%%%%
%% optional
%\supplementary{The following supporting information can be downloaded at:  \linksupplementary{s1}, Figure S1: title; Table S1: title; Video S1: title.}

% Only for journal Methods and Protocols:
% If you wish to submit a video article, please do so with any other supplementary material.
% \supplementary{The following supporting information can be downloaded at: \linksupplementary{s1}, Figure S1: title; Table S1: title; Video S1: title. A supporting video article is available at doi: link.}

% Only used for preprtints:
% \supplementary{The following supporting information can be downloaded at the website of this paper posted on \href{https://www.preprints.org/}{Preprints.org}.}

% Only for journal Hardware:
% If you wish to submit a video article, please do so with any other supplementary material.
% \supplementary{The following supporting information can be downloaded at: \linksupplementary{s1}, Figure S1: title; Table S1: title; Video S1: title.\vspace{6pt}\\
%\begin{tabularx}{\textwidth}{lll}
%\toprule
%\textbf{Name} & \textbf{Type} & \textbf{Description} \\
%\midrule
%S1 & Python script (.py) & Script of python source code used in XX \\
%S2 & Text (.txt) & Script of modelling code used to make Figure X \\
%S3 & Text (.txt) & Raw data from experiment X \\
%S4 & Video (.mp4) & Video demonstrating the hardware in use \\
%... & ... & ... \\
%\bottomrule
%\end{tabularx}
%}

%%%%%%%%%%%%%%%%%%%%%%%%%%%%%%%%%%%%%%%%%%
\authorcontributions{

Conceptualization, J.A., R.Á., A.R., and E.Y.; methodology, J.A., R.Á., A.R., and E.Y.; software, J.A. and E.Y.; validation, A.Z., J.T. and E.Y.; formal analysis, J.A. and E.Y.; investigation, J.A., R.Á., A.R., and E.Y.; resources, A.Z. and E.Y.; data curation, J.A. and E.Y.; writing---original draft preparation, J.A.; writing---review and editing, J.A., A.Z., J.T., R.Á., A.R., and E.Y.; visualization, J.A.; supervision, A.Z., J.T. and E.Y.; project administration, E.Y. All authors have read and agreed to the published version of the manuscript.
}

\funding{This research was funded by Regional Ministry of Economy, Knowledge, Enterprises, and University, Government of Andalusia (Spain) grant reference 1381382-F.}

\dataavailability{The dataset used during the current study are not publicly available due to privacy agreements with the jewelry companies that participated in this work. Access to the data is restricted in accordance with the confidentiality terms established by the collaborating stores.
}

\acknowledgments{We would like to express our gratitude for the cooperation and availability of the online jewelry stores that provided us access to their catalogs for the execution of the experiments in this study.

José Manuel Alcalde Llergo is a PhD student enrolled in the National PhD in Artificial Intelligence, XXXVIII cycle, course on Health and life sciences, organized by Università Campus Bio-Medico di Roma. He is also pursuing his doctorate in co-supervision at the Universidad de Córdoba (Spain), enrolled in its PhD program in Computation, Energy and Plasmas.}

\conflictsofinterest{The authors declare no conflicts of interest.}

%%%%%%%%%%%%%%%%%%%%%%%%%%%%%%%%%%%%%%%%%%
%\isPreprints{}{% This command is only used for ``preprints''.
\begin{adjustwidth}{-\extralength}{0cm}
%} % If the paper is ``preprints'', please uncomment this parenthesis.
%\printendnotes[custom] % Un-comment to print a list of endnotes

\reftitle{References}

% Please provide either the correct journal abbreviation (e.g. according to the “List of Title Word Abbreviations” http://www.issn.org/services/online-services/access-to-the-ltwa/) or the full name of the journal.
% Citations and References in Supplementary files are permitted provided that they also appear in the reference list here. 

%=====================================
% References, variant A: external bibliography

\bibliography{jewel_recog}
%=====================================
% \bibliography{your_external_BibTeX_file}

%=====================================
% References, variant B: internal bibliography
%=====================================

%\isPreprints{}{% This command is only used for ``preprints''.
\end{adjustwidth}
%} % If the paper is ``preprints'', please uncomment this parenthesis.
\end{document}